\documentclass{article}
\usepackage{spconf,amsmath,graphicx}
\usepackage{amssymb}
\usepackage{enumerate}
\usepackage{color}
\usepackage{booktabs}
\usepackage{multirow}
\usepackage{mathtools}
\usepackage{float} 

\title{Characterization of Brain Cortical Morphology Using Localized Topology-Encoding Graphs}

\name{Sevil~Maghsadhagh$^{1}$, Mousa~Shamsi$^{1}$, Anders Eklund$^{2,3}$, Hamid~Behjat$^{4}$ 
} 

\address{
\small $^{1}$ Department of Biomedical Engineering, Sahand University of Technology, Tabriz, Iran
\\ 
\small $^{2}$ Department of Biomedical Engineering, Link\"{o}ping University, Link\"{o}ping, Sweden
\\
\small $^{3}$ Department of Computer and Information Science, Link\"{o}ping University, Link\"{o}ping, Sweden
\\
\small $^{4}$ Department of Biomedical Engineering, Lund University, Lund, Sweden}

\begin{document}
%
\maketitle
\begin{abstract}
The human brain cortical layer has a convoluted morphology that is unique to each individual. Characterization of the cortical morphology is necessary in longitudinal studies of structural brain change,  as well as in discriminating individuals in health and disease. A method for encoding the cortical morphology in the form of a graph is presented. The design of graphs that encode the global cerebral hemisphere cortices as well as localized cortical regions is proposed. Spectral features of these graphs are then studied and proposed as descriptors of cortical morphology. As proof-of-concept of their applicability in characterizing cortical morphology, the descriptors are studied in the context of discriminating individuals based on their sex. 
\end{abstract}
\begin{keywords}
spectral graph theory, brain shape, cortical morphology
\end{keywords}
\vspace{-3mm}
\section{Introduction} 
\label{sec:intro}
The conventional approach for characterization of brain morphology and study of its changes is to quantify the volumes of a set of brain structures \cite{Decarli2005}. Cortical thickness measures are also popular means for characterizing morphology \cite{Fischl2000}. As volume and cortical thickness measures are incapable of capturing the full anatomical information, anatomical shape descriptors \cite{Gerardin2009, Shen2012} have been proposed to provide significant complementary informative representation of brain morphology. For example in \cite{Wachinger2015}, it was shown that shape descriptors of cortical and an ensemble of subcortical structures provides a powerful means to discriminate individuals based on their age, sex and neurodegenerative disorder. These shape descriptors use triangular surface mesh or tetrahedral volume tessellation constructions of brain structures, and exploit eigenfunctions of the LaplaceÐBeltrami operator \cite{Reuter2006}. 

Here we build on these works in several respects. Firstly, we use voxel-based graph designs. That is, we use the volumetric voxel representation of the 3D structure of the cortical ribbon, and construct a graph with vertices associated to individual voxels, and connectivities defined based on geodesic adjacencies. Previous designs of graphs with such encoding of the gray matter include, subject-specific designs of cerebral \cite{Behjat2013} and cerebellar \cite{Behjat2014} cortices, and template gray matter designs for a set of subjects \cite{Behjat2015}. In these studies, the graphs were exploited for analysis of fMRI data using graph signal processing \cite{Ortega2018} principles. The graphs were not explored for shape characterization. Secondly, we encode and exploit morphological information of an ensemble of localized cortical regions as opposed to using global shape descriptors of cortical structure. This is done by designing graphs that encode localized cortical regions. Thirdly, we propose the use of shape descriptors of graph spectral bands in contrast to using exact eigenvalues, and study how variations in topology of localized cortical regions is reflected across different spectral bands. 

\vspace{-3mm}
\section{Methods}
\label{sec:methods}

\subsection{Graphs and Their Spectra}
An undirected, unweighted graph $\mathcal{G}=(\mathcal{V},\mathcal{E},A)$ consists of a set $\mathcal{V}$ of $N_{g}:=|\mathcal{V}|$ vertices, a set $\mathcal{E}$ of edges  (i.e., pairs ($i,j$) where $i,j \in \mathcal{V}$), which can be fully described by an adajcency matrix $A$ with elements $A_{i,j}$ equal to $1$ if  $\text{if} \: (i,j) \in \mathcal{E}$, and $0$, if otherwise. Using $A$, the graph's diagonal, degree matrix $D$ is defined with elements $ D_{i,i}=\sum_{j} A_{i,j}$, and the graph's normalized Laplacian matrix $L$ is defined as 
\begin{align} 
L & =  I - D^{-1/2} A D^{-1/2}.
\label{eq:L}
\end{align} 

\noindent
Since $L$ is symmetric and positive semi-definite, it can be diagonalized as $
L =  \Sigma \Lambda {\Sigma}^{T},
$ where $ \Sigma = [\chi_{_1} | \chi_{_2} | \cdots | \chi_{_{N_{g}}}],
$ is an orthonormal matrix containing a set of $N_g$ eigenvectors $\{\chi_{_k}\}_{k=1}^{N_{g}}$, and $\Lambda$ is a diagonal matrix whose entries equal the associated real, non-negative eigenvalues that define the graph spectrum $\mathcal{S}$ as
\begin{equation}
\label{eq:graphSpectrum}
\mathcal{S} = \text{diag}(\Lambda) = \{ 0 = \lambda_{1} \le \lambda_{2} \le \cdots \le  \lambda_{N_{g}} \le 2 \}.
\end{equation}
Unlike classical Euclidean domain spectrum, each graph has a unique definition of spectrum, with a unique range $[0,  \lambda_{N_{g}}]$ and a unique set of irregularly spaced eigenvalues with possibility of multiplicity greater than one. 
 
\subsection{Cortical Morphology Encoding Graphs}

\subsubsection{Global Cerebral Hemisphere Cortex Graphs}
For each hemisphere, a graph that encodes the cortical topology is designed. Cortical ribbons \cite{Dale1999, Fischl2000} constructed using the FreeSurfer software package \cite{Fischl2012} are exploited. The voxels within the cortical ribbon are treated as graph vertices. The graph edges are defined based on 26-neighborhood connectivity of voxels in 3D space. Two vertices are connected through an edge if they lie within each-other's 26-neighborhood. Due to limited voxel resolution, edges derived merely based on Euclidean adjacency may include some spurious connections that are not anatomically justifiable, for instance, at touching banks of sulci where the two banks are adjacent in a Euclidean sense but anatomically non-adjacent. By exploiting pial surface extractions, such anatomically unjustifiable connections, i.e. graph edges, are pruned out. No weight is assigned to the edges. The resulting left and right hemisphere graphs are denoted $\mathcal{G}^{(l)}$Êand $\mathcal{G}^{(r)}$, respectively, and their normalized Laplacian matrices are denoted $L^{(l)}$ and $L^{(r)}$, respectively.
 
\subsubsection{Cortical Parcellation}
\label{sec:parcellation}
Each hemisphere is parcellated into a set of regions of approximately equal volume.   For satisfying equality of regional volumes, the number of regions may slightly vary between the two hemispheres depending on the level of asymmetry in the volume of the two hemispheres. The parcellation is performed using spectral clustering \cite{Luxburg2007} by exploiting the eigenvectors of $L^{(l)}$ and $L^{(r)}$. For instance, the parcellation of the left hemisphere into $N$ regions is obtained as follows. Initially, a set of vectors are defined as 
\begin{align}
y_{i} 
& = [\chi_{_1}[i],\chi_{_2}[i], \ldots,\chi_{_N}[i]],  \quad i=1, \ldots, N_g^{(l)}, 
\end{align}
where $\{\chi_{_n}\}_{n=1}^{N}$ denote the first $N$ eigenvectors of $L^{(l)}$ and $N_g^{(l)}$ denotes the number of vertices of $\mathcal{G}^{(l)}$.
The data vectors $\{y_{i}\}_{i=1}^{N_{g}^{(l)}}$ are then clustered with the k-means algorithm in to $N$ clusters $\{C_{j} \subset \{1,2,...,N_g^{(l)}\} \}_{j=1}^{N}$, where $C_{1}~\cup~C_{2}~\cup~\cdots~C_{N} = \{1,2,...,N_g^{(l)}\}$. Voxels associated to each vertex subset $C_{i}$ are then treated as a region, resulting in $N$ localized cortical regions within the hemisphere.

\subsubsection{Localized Cerebral Hemisphere Cortex Graphs}
A graph is designed for each localized cortical region. The vertex set of the graph associated to each cortical region $i$ consists of voxels that lie within the associated cortical parcel $C_{i}$. The graph's edges are defined based on the same neigbouhood connectivity principle as that explained in constructing $\mathcal{G}^{(l)}$ and $\mathcal{G}^{(r)}$. In practice, the adjacency matrices of the local graphs are extracted from the adjacency matrix of the global graph of the associated hemisphere.    

\begin{figure}[tbp] 
   \centering
   \includegraphics[width=0.45\textwidth]{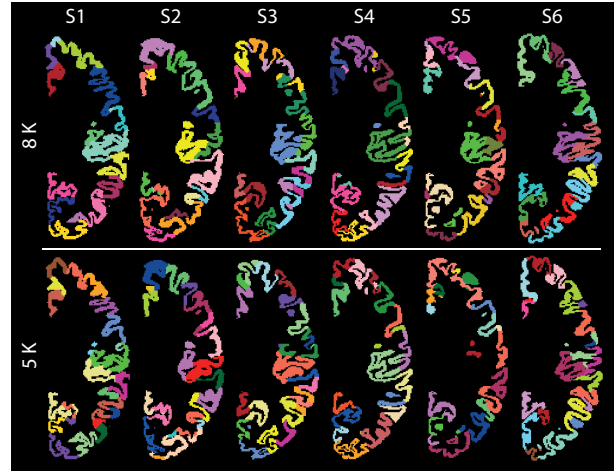}
   \caption{Parcellated right hemisphere of six subjects from the HCP database, one subject per column, at two resolutions; 40Ê$\pm$ 2 and 63 $\pm$ 3 parcels in the top and bottom rows, respectively. Slices are shown along the same MNI coordinate, at 1 mm$^{3}$ voxel resolution. Volumes of cortical regions in each hemisphere are equal, and are approximately equal across subjects; top row: $~$Ê8000 (8K) voxels, bottom row: $~$ 5000 (5K) voxels. The first three subjects are females and the second three are males.}
   \label{fig:regions} 
\vspace{-3mm}
\end{figure}

\section{Results}
The analysis is performed on 30 subjects from the Human Connectome Project \cite{HCP} database, consisting of 15 female and 15 male subjects. Fig.~\ref{fig:regions} shows the cortical structure of the right hemisphere of 6 subjects, where the first three subjects are females and the second three are males. The figure also shows example cortical parcellations using the scheme described in Section~\ref{sec:parcellation}. The local graphs associated to the cortical regions in the top row and the bottom row have $~$5 K and 8 K vertices, respectively. In the following, results will be presented on local graphs of size 5K to 10K vertices.

\begin{figure}[t] 
   \centering
   \includegraphics[width=0.47\textwidth]{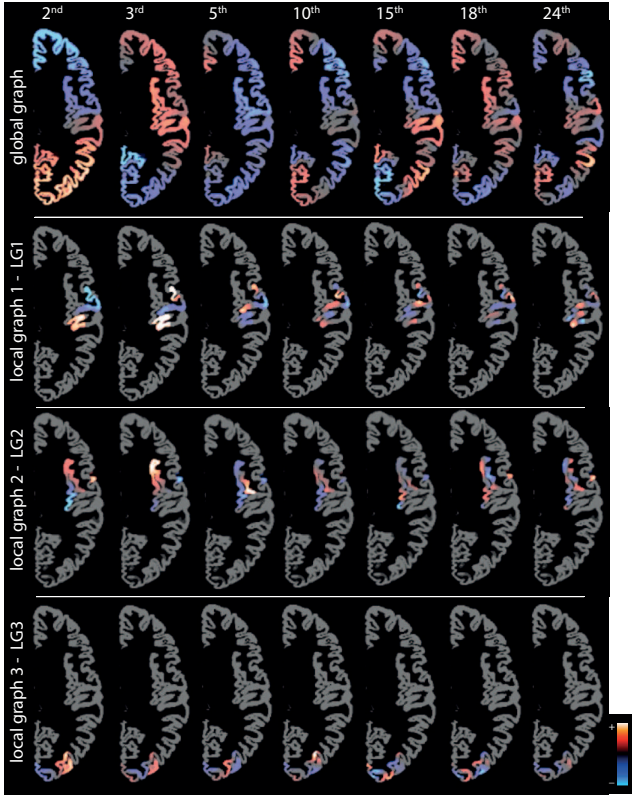} 
   \caption{Eigenvectors of the global graph and three local graphs associated to the right hemisphere of subject S1, cf. Fig.~\ref{fig:regions}, are displayed. Note that the eigenvectors are defined in 3D space whereas only a single axial slice is shown, which limits showing the global spatial variation.}
   \label{fig:eigenvectors}
   \vspace{-3mm}
\end{figure}

\subsection{Spectra of local and global cortical graphs}
To provide insight on the nature of the proposed cortical graphs, it is intuitive to depict several of the graphs' eigenvectors. Fig.~\ref{fig:eigenvectors} shows eigenvectors of the Laplacian matrices of three local cortical graphs and the associated global hemisphere graph. Eigenvectors associated to smaller eigenvalues represent slower spatial harmonics, whereas those associated to higher eigenvalues, encode more subtle spatial patterns. The eigenvectors of the local graph better capture localized morphological variations, whereas the corresponding vectors in the global graph capture more global topological variations. The constructed local graphs are of size 8 K vertices, and they thus, have 8 K eigenvectors. Yet, it is interesting to observe that the initial eigenvectors capture noticeable cortical variation. In the analysis that follows, we focus on interpreting and processing information associated only to the lower-end of the graph spectra.

\begin{figure*}[h] 
   \centering
   \includegraphics[width=0.95\textwidth]{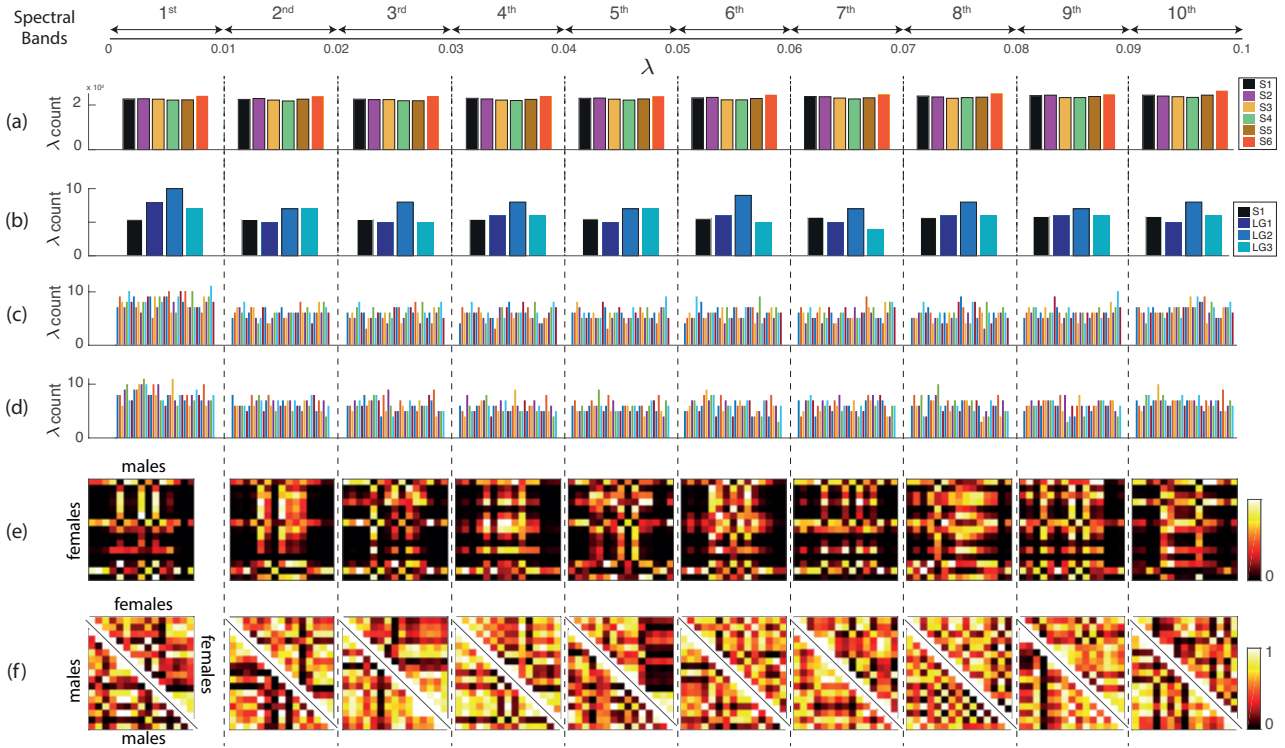} 
    \caption{(a)-(d) Distribution of the eigenvalues in the lower-end spectra of cortical graphs. The graphs' spectral range $[0 - 0.1]$ is split in to ten bands. (a) Global graphs associated to the six right hemispheres shown in Fig.~\ref{fig:regions}. (b) Global graph and the three local graphs associated to the hemisphere shown in Fig.~\ref{fig:eigenvectors}. The black bars in (b) are obtained by dividing the black bars in (a) by the number of local graphs in the hemisphere, $N = 42$. (c) 42 local graphs associated to the hemisphere shown in Fig.~\ref{fig:eigenvectors}. (d) The same as in (c) but for the subject's left hemisphere, $N = 41$. (e) Uncorrected p-values resulting from Wilcoxon rank-sum test on different spectral band features between subjects of opposite sex. (f) The same as in (e) but between subjects of the same sex. In (b)-(f), local graphs have size 8K; see Tables~\ref{tab:pvals-pairwise} and \ref{tab:pvals-mean-features} for statistical results on other local graph sizes.}
   \label{fig:merged}
\end{figure*}

Fig.~\ref{fig:merged}(a), shows the distribution of eigenvalues in the lower-end spectra of global graphs associated to the six hemispheres shown in Fig.~\ref{fig:regions}. The spectra show minute variation across ten spectral bands. The choice of ten spectral bans was heuristic, but the analysis can be readily extended to coarser or finer spectral bands. Fig.~\ref{fig:merged}(b) compares the distribution of the lower-end spectra of the global graph and the three local graphs shown in Fig.~\ref{fig:eigenvectors}. The variation observed in the spectral distribution of the local graphs can be interpreted as reflecting variations in local cortical morphology in the associated three cortical parcels. 

Fig.~\ref{fig:merged}(c) shows the distribution of lower-end spectra of all local graphs of the hemisphere shown Fig.~\ref{fig:eigenvectors}; Fig.~\ref{fig:merged}(d) shows the distributions for the subject's left hemisphere. In both hemispheres, a considerable level of variation is observed across spectral bands and local graphs. The distributions can be seen as a decomposition of the single distribution of the global hemisphere graph in to multiple distributions; i.e., splitting the height of the black bars in Fig.~\ref{fig:merged}(a) in to the set of distributions shown in Fig.~\ref{fig:merged}(c). The extent of information observed in this implicit decomposition process suggests the potential benefit of local cortical graph spectra as a powerful means to discriminate cortical morphology across subjects. 

   \begin{table}[b]
   \centering
   \resizebox{0.36\textwidth}{!}
   {
   \begin{tabular}{@{} r @{\hspace{1.5\tabcolsep}} r @{\hspace{1.5\tabcolsep}} r @{\hspace{1.5\tabcolsep}} r @{\hspace{1.5\tabcolsep}} r @{\hspace{1.5\tabcolsep}} r @{\hspace{1.5\tabcolsep}}r @{\hspace{1.5\tabcolsep}} r @{\hspace{1.5\tabcolsep}} r @{\hspace{1.5\tabcolsep}} r @{\hspace{1.5\tabcolsep}} r @{\hspace{1.5\tabcolsep}} r @{\hspace{1.5\tabcolsep}} c @{}} 
      \toprule
      & \multicolumn{10}{c}{Spectral Bands} \\
      \cmidrule{3-12} 
      & &  1$^{\text{st}}$ & {2$^{\text{nd}}$} & {3$^{\text{rd}}$} & {4$^{\text{th}}$} & {5$^{\text{th}}$} & {6$^{\text{th}}$} & {7$^{\text{th}}$} & {8$^{\text{th}}$} & {9$^{\text{th}}$} & {10$^{\text{th}}$} & {average}\\
      \midrule
      \multicolumn{1}{c}{\multirow{2}{*}{5K}} & OS & 62   & 48   & 48  &  43  &  46  &  46  &  38   & 40  &  43   & 42 & 46\\
  & SS  & 12   &  2   &  3  &  10  &  10   &  6   &  8  &   1  &   8   & 11 & 7 \\
    \cmidrule{1-13}
\multicolumn{1}{c}{\multirow{2}{*}{6K}} & OS  & 62 &   53  &  43  &  42  &  46 &   45  &  32  &  39 &   35   & 40 & 44 \\
  & SS   & 2    & 5   &  6  &   4   &  2  &   6   &  5    & 8  &   8    & 8& 5\\
\cmidrule{1-13}
\multicolumn{1}{c}{\multirow{2}{*}{7K}} & OS & 56  &  54  &  50  &  41  &  48 &   45 &   41   & 39 &   40  &  53 & 47\\
 & SS &  5 &    3  &   3  &   0  &   0 &    1  &   1   &  4  &   0  &   4 & 2\\
\cmidrule{1-13}
\multicolumn{1}{c}{\multirow{2}{*}{8K}} & OS & 55   & 48   & 51   & 47  &  40   & 40   & 45   & 27  &  39 &   44 & 44\\
 & SS &     6  &  11 &    4  &   1 &   10  &   2  &   1  &   0  &   4  &   4 & 4\\
\cmidrule{1-13}
\multicolumn{1}{c}{\multirow{2}{*}{9K}} & OS &  53  &  52  &  51 &   49  &  50   & 34   & 40  &  38   & 46   & 45 & 46\\
& SS &     3   &  3   &  0   & 10 &  2 &    4   &  7   &  5    & 9   & 10 & 5\\
\cmidrule{1-13}
\multicolumn{1}{c}{\multirow{2}{*}{10K}} & OS & 49 &   46 &   53 &   35 &   49 &   42 &   28 &   38 &   48 &   43 & 43 \\
& SS &    0   &  3 &    4  &   4  &   0  &   0  &   5  &   3   &  8  &   2 & 3\\
\bottomrule
   \end{tabular}
    } 
    \caption{Percentage of Wilcoxon rank-sum tests between opposite sex (OS) and same sex (SS) subjects with uncorrected p-values $< 0.05$; all values are in $\%$; see the description in text for details.}   \label{tab:pvals-pairwise}
   \vspace{-3mm}
  \end{table}

\subsection{Discriminating sex based on cortical topology}

Variations in the cortical morphology of male and female subjects has been suggested in many studies, see for example \cite{Ritchie2018}. The potential of cortical descriptors of graph spectral band, i.e., $\lambda$ counts as shown in Fig.~\ref{fig:merged}(a)-(d), was investigated for the task of discriminating female and male subjects. A Wilcoxon rank-sum test was performed between every male and female subject, using the $\lambda$ counts in each spectral band. For the test at the $n$th spectral band, the sample set of each subject consists of $\lambda$ counts of the $n$th spectral band of all local graphs across the two hemispheres. Figs.~\ref{fig:merged}(e)-(f) show the resulting p-values (uncorrected) from the tests, across different spectral bands and different local graph sizes. The tests between subjects of opposite sex exhibit smaller p-values, as can be seen by darker values across more pixels in subplots in Fig.~\ref{fig:merged}(e), in contrast to higher p-values observed in tests between subject of same sex, see subplots in Fig.~\ref{fig:merged}(f). This observation is consistent across spectral bands. The analysis was performed on graphs of varying size; the percentage of the tests that led to p-values $<$ 0.05 is reported in Table~\ref{tab:pvals-pairwise}. The average performance across spectral bands for each graph size is shown in the last column, which shows a large difference between the proportion of significant tests between same sex and opposite sex subjects.    

\begin{table}[h]
\vspace{0mm} 
\resizebox{0.49 \textwidth}{!}{
   \centering
   \begin{tabular}{@{} c @{\hspace{1.3\tabcolsep}} r @{\hspace{1.3\tabcolsep}}r @{\hspace{1.3\tabcolsep}} r @{\hspace{1.3\tabcolsep}}r @{\hspace{1.3\tabcolsep}}r @{\hspace{1.3\tabcolsep}}r @{\hspace{1.3\tabcolsep}} c  @{}} 
      \toprule
      \multicolumn{1}{c}{\multirow{2}{*}{SB}} & \multicolumn{6}{c}{Local Graphs} & \multicolumn{1}{c}{\multirow{1}{*}{Global}}\\
      \cmidrule{2-7} 
      & \multicolumn{1}{c}{5K} & \multicolumn{1}{c}{6K} & \multicolumn{1}{c}{7K} & \multicolumn{1}{c}{8K} & \multicolumn{1}{c}{9K} & \multicolumn{1}{c}{10K} & Graphs\\
      \midrule
      1$^{\text{st}}$     
      & 1.5 {$\times 10^{-6}$}
      & 3.6 {$\times 10^{-6}$}  
      & 3.1 {$\times 10^{-6}$} 
      & 1.3 {$\times 10^{-5}$} 
      & 6.7 {$\times 10^{-5}$}
      & 4.7 {$\times 10^{-6}$} 
      & 1.8 {$\times 10^{-4}$} \\
      2$^{\text{nd}}$      
      & 1.5 {$\times 10^{-4}$}        
      & 2.2 {$\times 10^{-4}$} 
      & 3.4 {$\times 10^{-6}$} 
      & 3.6 {$\times 10^{-5}$} 
      & 5.6 {$\times 10^{-6}$}
      & 9.8 {$\times 10^{-5}$} 
      & 3.5 {$\times 10^{-4}$} \\
      3$^{\text{rd}}$   
      & 1.3 {$\times 10^{-4}$} 
      & 4.2 {$\times 10^{-6}$} 
      & 1.1 {$\times 10^{-5}$}  
      & 5.2 {$\times 10^{-5}$} 
      & ~3 {$\times 10^{-5}$} 
      & 1.5 {$\times 10^{-6}$} 
      & 2.9 {$\times 10^{-4}$}  \\
      4$^{\text{th}}$       
      & 1.4 {$\times 10^{-4}$} 
      & 5.9 {$\times 10^{-5}$} 
      & 2.5 {$\times 10^{-5}$} 
      & ~1 {$\times 10^{-6}$}
      & 1.8 {$\times 10^{-4}$}
      & 2.1 {$\times 10^{-5}$}
      & 1.2 {$\times 10^{-3}$}
 \\
      5$^{\text{th}}$  
& 1.5 {$\times 10^{-4}$}
& 3.9 {$\times 10^{-6}$}
& 6.2 {$\times 10^{-6}$}
& 1.9 {$\times 10^{-3}$}
& 2.4 {$\times 10^{-6}$}
& 2.9 {$\times 10^{-5}$}
& 3.6 {$\times 10^{-4}$}
\\
         6$^{\text{th}}$  
& 5.3 {$\times 10^{-4}$}
& 7.9 {$\times 10^{-5}$}
& 1.9 {$\times 10^{-4}$}
& 4.9 {$\times 10^{-5}$}
& 1.7 {$\times 10^{-5}$}
& 1.1 {$\times 10^{-6}$}
& 4.9 {$\times 10^{-3}$}
 \\
            7$^{\text{th}}$  
& 6.1 {$\times 10^{-5}$}
& 2.4 {$\times 10^{-4}$}
& 6.5 {$\times 10^{-5}$}
& 1.8 {$\times 10^{-5}$}
& ~4 {$\times 10^{-4}$}
& 7.9 {$\times 10^{-3}$}
& 8.8 {$\times 10^{-3}$}
\\
               8$^{\text{th}}$  
& 2.6 {$\times 10^{-4}$}
& 1.2 {$\times 10^{-3}$}
& 1.6 {$\times 10^{-5}$}
& 6.9 {$\times 10^{-3}$}
& 1.3 {$\times 10^{-4}$}
& 2.5 {$\times 10^{-5}$}
& 4.4 {$\times 10^{-3}$}
 \\
                  9$^{\text{th}}$  
& 5.4 {$\times 10^{-4}$}
& 3.1 {$\times 10^{-5}$}
& 4.1 {$\times 10^{-4}$}
& 6.1 {$\times 10^{-5}$}
& 1.4 {$\times 10^{-5}$}
& 2.5 {$\times 10^{-6}$}
& 4.2 {$\times 10^{-3}$}
 \\
                     10$^{\text{th}}$  
& 1.8 {$\times 10^{-4}$}
& 1.5 {$\times 10^{-5}$}
& 1.3 {$\times 10^{-6}$}
& 1.3 {$\times 10^{-4}$}
& 1.1 {$\times 10^{-5}$}
& 2.3 {$\times 10^{-5}$}
& ~3 {$\times 10^{-2}$}
 \\
      \bottomrule
   \end{tabular}
} 
\caption{Uncorrected p-values resulting from Wilcoxon rank-sum tests between the group of male and female subjects using mean spectral features of different spectral bands (SB); see the description in text for details.}
\label{tab:pvals-mean-features}
\vspace{-3mm}
\end{table}

Wilcoxon rank-sum test analysis was then performed on the group of male and females subjects. The tests were performed on the mean spectral band value of each group; i.e. the $\lambda$ count was averaged over all local graphs of each hemisphere, resulting in a two values for each subject at each spectral band and for each graph size. The values were pulled together for the 15 subjects of each sex, thus resulting in samples of size 30. Resulting p-values (uncorrected) are shown in Table~\ref{tab:pvals-mean-features}. The last column shows p-values obtained using $\lambda$ counts of the left and right hemisphere global graphs, with sample sizes of 30 for each sex. With the exception of the 5th and 8th spectral bands of the 8K design, all tests on local graphs lead to lower p-values relative to the associated test on global graphs, suggesting the superiority of local graphs over global graphs in discriminative encoding of cortical topology.

\vspace{-4mm}

\section{Conclusions}
The design of cerebral cortical graphs, consisting of global hemisphere graphs and localized cortical graphs, was presented. Global hemisphere graphs encode the global topology cerebral hemisphere cortices, whereas local cortical graphs capture more subtle localized variations in cortical morphology. The set of spectra of local cortical graphs can be seen as an implicit decomposition of the single spectrum of the associated global hemisphere graph. Experimental results suggest the benefit of spectral features of cortical graphs as a powerful means for discriminative characterization of cortical morphology across individuals. In future work, we will focus on testing the proposed cortical graph features on a larger cohort of healthy as well as patient subjects; in particular,  characterization and early detection of changes in cortical morphology that arise in Alzheimer's disease \cite{Falahati2014} will be explored. The proposed local cortical graphs can also be found applicable for enhanced graph spectral processing of functional MRI data \cite{Behjat2015, HuangBolton2018} using novel principles \cite{Behjat2016, Stankovic2018} from the recently emerged field of graph signal processing \cite{Ortega2018}.


\clearpage
\bibliographystyle{IEEEbib}
\bibliography{hbehjat_bibliography}

\begin{thebibliography}{10}

\bibitem{Decarli2005}
C.~DeCarli, J.~Massaro, D.~Harvey, J.~Hald, M.~Tullberg, R.~Au, A.~Beiser,
  R.~D'Agostino, and P.A. Wolf,
\newblock ``Measures of brain morphology and infarction in the framingham heart
  study: establishing what is normal,''
\newblock {\em Neurobiology of aging}, vol. 26, no. 4, pp. 491--510, 2005.

\bibitem{Fischl2000}
B.~Fischl and A.M. Dale,
\newblock ``Measuring the thickness of the human cerebral cortex from magnetic
  resonance images,''
\newblock {\em Proc. Natl Acad. Sci.}, vol. 97, no. 20, pp. 11050--11055, 2000.

\bibitem{Gerardin2009}
E.~Gerardin, G.~Ch{\'e}telat, M.~Chupin, R.~Cuingnet, B.~Desgranges, H.~Kim,
  M.~Niethammer, B.~Dubois, S.~Leh{\'e}ricy, L.~Garnero, et~al.,
\newblock ``Multidimensional classification of hippocampal shape features
  discriminates alzheimer's disease and mild cognitive impairment from normal
  aging,''
\newblock {\em Neuroimage}, vol. 47, no. 4, pp. 1476--1486, 2009.

\bibitem{Shen2012}
K.~Shen, J.~Fripp, F.~M{\'e}riaudeau, G.~Ch{\'e}telat, O.~Salvado, P.~Bourgeat,
  Alzheimer's Disease~Neuroimaging Initiative, et~al.,
\newblock ``Detecting global and local hippocampal shape changes in alzheimer's
  disease using statistical shape models,''
\newblock {\em Neuroimage}, vol. 59, no. 3, pp. 2155--2166, 2012.

\bibitem{Wachinger2015}
C.~Wachinger, P.~Golland, W.~Kremen, B.~Fischl, M.~Reuter, Alzheimer's
  Disease~Neuroimaging Initiative, et~al.,
\newblock ``Brainprint: a discriminative characterization of brain
  morphology,''
\newblock {\em Neuroimage}, vol. 109, pp. 232--248, 2015.

\bibitem{Reuter2006}
M.~Reuter, F-E. Wolter, and N.~Peinecke,
\newblock ``Laplace--beltrami spectra as Ôshape-dnaÕof surfaces and solids,''
\newblock {\em Computer-Aided Design}, vol. 38, no. 4, pp. 342--366, 2006.

\bibitem{Behjat2013}
H.~Behjat, N.~Leonardi, and D.~{Van De Ville},
\newblock ``Statistical parametric mapping of functional {MRI} data using
  wavelets adapted to the cerebral cortex,''
\newblock in {\em Proc. IEEE Int. Symp. Biomed. Imaging}, 2013, pp. 1070--1073.

\bibitem{Behjat2014}
H.~Behjat, N.~Leonardi, L.~S\"{o}rnmo, and D.~{Van De Ville},
\newblock ``Canonical cerebellar graph wavelets and their application to {fMRI}
  activation mapping,''
\newblock in {\em Proc. IEEE Int. Conf. Eng. Med. Biol. Soc.}, 2014, pp.
  1039--1042.

\bibitem{Behjat2015}
H.~Behjat, N.~Leonardi, L.~S\"{o}rnmo, and D.~{Van De Ville},
\newblock ``Anatomically-adapted graph wavelets for improved group-level {fMRI}
  activation mapping,''
\newblock {\em Neuroimage}, vol. 123, pp. 185--199, 2015.

\bibitem{Ortega2018}
A.~Ortega, P.~Frossard, J.~Kova{\v{c}}evi{\'c}, J.~M.~F. Moura, and
  P.~Vandergheynst,
\newblock ``Graph signal processing: Overview, challenges, and applications,''
\newblock {\em Proc. IEEE}, vol. 106, no. 5, pp. 808--828, May 2018.

\bibitem{Dale1999}
A.M. Dale, B.~Fischl, and M.I. Sereno,
\newblock ``Cortical surface-based analysis: I. segmentation and surface
  reconstruction,''
\newblock {\em Neuroimage}, vol. 9, no. 2, pp. 179--194, 1999.

\bibitem{Fischl2012}
B.~Fischl,
\newblock ``Freesurfer,''
\newblock {\em Neuroimage}, vol. 62, no. 2, pp. 774--781, 2012.

\bibitem{Luxburg2007}
U.~Von~Luxburg,
\newblock ``A tutorial on spectral clustering,''
\newblock {\em Stat. Comput.}, vol. 17, no. 4, pp. 395--416, 2007.

\bibitem{HCP}
D.C. {Van Essen}, S.M. Smith, D.M. Barch, T.E. Behrens, E.~Yacoub, K.~Ugurbil,
  and WU-Minn~HCP Consortium.,
\newblock ``The {WU-Minn} human connectome project: an overview.,''
\newblock {\em Neuroimage}, vol. 80, pp. 62--79, 2013.

\bibitem{Ritchie2018}
S.J. Ritchie, S.R. Cox, X.~Shen, M.V. Lombardo, L.M. Reus, C.~Alloza, M.A.
  Harris, H.L. Alderson, S.~Hunter, E.~Neilson, et~al.,
\newblock ``Sex differences in the adult human brain: evidence from 5216 uk
  biobank participants,''
\newblock {\em Cerebral Cortex}, vol. 28, no. 8, pp. 2959--2975, 2018.

\bibitem{Falahati2014}
F.~Falahati, E.~Westman, and A.~Simmons,
\newblock ``Multivariate data analysis and machine learning in alzheimer's
  disease with a focus on structural magnetic resonance imaging,''
\newblock {\em Journal of Alzheimer's Disease}, vol. 41, no. 3, pp. 685--708,
  2014.

\bibitem{HuangBolton2018}
W.~Huang, T.~A.~W. Bolton, J.~D. Medaglia, D.~S. Bassett, A.~Ribeiro, and
  D.~Van~De Ville,
\newblock ``A graph signal processing perspective on functional brain
  imaging,''
\newblock {\em Proceedings of the IEEE}, vol. 106, no. 5, pp. 868--885, May
  2018.

\bibitem{Behjat2016}
H.~Behjat, U.~Richter, D.~{Van De Ville}, and L.~S\"{o}rnmo,
\newblock ``Signal-adapted tight frames on graphs.,''
\newblock {\em IEEE Trans. Signal Process.}, vol. 64, no. 22, pp. 6017--6029,
  2016.

\bibitem{Stankovic2018}
L.~Stankovi{\'c}, E.~Sejdi{\'c}, and M.~Dakovi{\'c},
\newblock ``Reduced interference vertex-frequency distributions,''
\newblock {\em IEEE Signal Processing Letters}, vol. 25, no. 9, pp. 1393--1397,
  2018.

\end{thebibliography}

\end{document}